\documentclass[sigconf]{acmart}
\AtBeginDocument{%
  \providecommand\BibTeX{{%
    \normalfont B\kern-0.5em{\scshape i\kern-0.25em b}\kern-0.8em\TeX}}}

\copyrightyear{2021}
\acmYear{2021}
\setcopyright{rightsretained}
\acmConference[KDD '21]{Proceedings of the 27th ACM SIGKDD
Conference on Knowledge Discovery and Data Mining}{August 14--18,
2021}{Virtual Event, Singapore}
\acmBooktitle{Proceedings of the 27th ACM SIGKDD Conference on
Knowledge Discovery and Data Mining (KDD '21), August 14--18, 2021,
Virtual Event, Singapore}
\acmPrice{}
\acmISBN{978-1-4503-8332-5/21/08}
\acmDOI{10.1145/3447548.3467371}

\usepackage{amsthm}
\theoremstyle{definition}

\usepackage{amsmath,amsfonts,bm}

\def\eqref#1{equation~\ref{#1}}
\def\1{\bm{1}}

\def\vtheta{{\bm{\theta}}}

\def\vh{{\bm{h}}}

\def\vw{{\bm{w}}}
\def\vx{{\bm{x}}}
\def\vy{{\bm{y}}}

\def\mA{{\bm{A}}}

\def\mD{{\bm{D}}}

\def\mL{{\bm{L}}}

\def\mW{{\bm{W}}}

\DeclareMathAlphabet{\mathsfit}{\encodingdefault}{\sfdefault}{m}{sl}
\SetMathAlphabet{\mathsfit}{bold}{\encodingdefault}{\sfdefault}{bx}{n}
\newcommand{\tens}[1]{\bm{\mathsfit{#1}}}

\def\tX{{\tens{X}}}
\def\tY{{\tens{Y}}}

\usepackage{multirow}

\usepackage{hyperref}
\hypersetup{
    colorlinks=true,
    citecolor=blue
}

\usepackage{float}
\usepackage{url}
\usepackage{booktabs}
\usepackage{graphicx}
\usepackage{algorithm} 
\usepackage{algpseudocode}
\usepackage{varwidth}% http://ctan.org/pkg/varwidth
\usepackage{multicol}
\usepackage{multirow}
\usepackage{caption}
\usepackage{subcaption}
\usepackage{wrapfig}

\newcommand{\fref}[1]{Figure~\ref{#1}}
\newcommand{\eref}[1]{Equation~\ref{#1}}
\newcommand{\aref}[1]{Algorithm~\ref{#1}}
\newcommand{\tref}[1]{Table~\ref{#1}}

\newcommand{\update}[1]{{#1}}

\def\modelfullname{{Cross-Node Federated Graph Neural Network}}
\def\modelshortname{{CNFGNN}}

\begin{document}
\fancyhead{}

\title{Cross-Node Federated Graph Neural Network for Spatio-Temporal Data Modeling}

\author{Chuizheng Meng}
\affiliation{%
  \institution{University of Southern California}
  \city{Los Angeles}
  \state{California}
  \country{USA}
}
\email{chuizhem@usc.edu}

\author{Sirisha Rambhatla}
\affiliation{%
  \institution{University of Southern California}
  \city{Los Angeles}
  \state{California}
  \country{USA}
}
\email{sirishar@usc.edu}

\author{Yan Liu}
\affiliation{%
  \institution{University of Southern California}
  \city{Los Angeles}
  \state{California}
  \country{USA}
}
\email{yanliu.cs@usc.edu}

\renewcommand{\shortauthors}{Trovato and Tobin, et al.}

\begin{abstract}
Vast amount of data generated from networks of sensors, wearables, and the Internet of Things (IoT) devices underscores the need for advanced modeling techniques that leverage the spatio-temporal structure of decentralized data due to the need for edge computation and licensing (data access) issues. While federated learning (FL) has emerged as a framework for model training without requiring direct data sharing and exchange, effectively modeling the complex spatio-temporal dependencies to improve forecasting capabilities still remains an open problem. On the other hand, state-of-the-art spatio-temporal forecasting models assume unfettered access to the data, neglecting constraints on data sharing. To bridge this gap, we propose a federated spatio-temporal model -- \modelfullname~(\modelshortname) -- which explicitly encodes the underlying graph structure using graph neural network (GNN)-based architecture under the constraint of cross-node federated learning, which requires that data in a network of nodes is generated locally on each node and remains decentralized. \modelshortname ~operates by disentangling the temporal dynamics modeling on devices and spatial dynamics on the server, utilizing alternating optimization to reduce the communication cost, facilitating computations on the edge devices. Experiments on the traffic flow forecasting task show that \modelshortname~achieves the best forecasting performance in both transductive and inductive learning settings with no extra computation cost on  edge devices, while incurring modest communication cost.
\end{abstract}

\begin{CCSXML}
<ccs2012>
   <concept>
       <concept_id>10002951.10003227.10003236.10003238</concept_id>
       <concept_desc>Information systems~Sensor networks</concept_desc>
       <concept_significance>500</concept_significance>
       </concept>
   <concept>
       <concept_id>10002951.10003227.10003351</concept_id>
       <concept_desc>Information systems~Data mining</concept_desc>
       <concept_significance>300</concept_significance>
       </concept>
   <concept>
       <concept_id>10010147.10010257.10010293.10010294</concept_id>
       <concept_desc>Computing methodologies~Neural networks</concept_desc>
       <concept_significance>500</concept_significance>
       </concept>
 </ccs2012>
\end{CCSXML}

\ccsdesc[500]{Information systems~Sensor networks}
\ccsdesc[300]{Information systems~Data mining}
\ccsdesc[500]{Computing methodologies~Neural networks}

% \keywords{datasets, neural networks, gaze detection, text tagging}
\keywords{Federated Learning; Graph Neural Network; Spatio-Temporal Data Modeling}


\maketitle

\section{Introduction}

%v1
%Modeling the dynamics of spatio-temporal data generated from networks of nodes (e.g. sensors, wearable devices and the Internet of Things (IoT) devices) is critical for various applications including traffic flow prediction~\citep{li2018dcrnn_traffic,yu2018spatio}, forecasting~\citep{seo2019physics,azencot2020forecasting}, and user activity detection~\citep{yan2018spatial,liu2020disentangling}. Since spatio-temporal data is usually distributed on nodes belonging to different owners and contains sensitive information, it cannot be easily shared or transmitted among nodes due to privacy concerns or data access restrictions due to licensing issues. Nevertheless, most existing works on spatio-temporal dynamics modeling~\citep{battaglia2016interaction,kipf2018neural,battaglia2018relational} assume that the model is trained with centralized data gathered from all devices. Recent developments of federated learning (FL)~\citep{kairouz2019advances} provides a solution for training a model with decentralized data on multiple devices. However, existing works in FL either ignore the spatio-temporal dependencies of decentralized data~\citep{mcmahan2017communication,li2018federated,karimireddy2019scaffold} or model it by implicitly imposing the graph structure~\citep{smith2017federated} and cannot operate in settings where only a fraction of devices are observed during training (\emph{inductive learning setting}).

%v2 (sirisha)
Modeling the dynamics of spatio-temporal data generated from networks of edge devices or nodes (e.g. sensors, wearable devices and the Internet of Things (IoT) devices) is critical for various applications including traffic flow prediction~\citep{li2018dcrnn_traffic,yu2018spatio}, forecasting~\citep{seo2019physics,azencot2020forecasting}, and user activity detection~\citep{yan2018spatial,liu2020disentangling}. While existing works on spatio-temporal dynamics modeling~\citep{battaglia2016interaction,kipf2018neural,battaglia2018relational} assume that the model is trained with centralized data gathered from all devices, the volume of data generated at these edge devices precludes the use of such centralized data processing, and calls for decentralized processing where computations on the edge can lead to significant gains in improving the latency. In addition, in case of spatio-temporal forecasting, the edge devices need to leverage the complex inter-dependencies to improve the prediction performance. Moreover, with increasing concerns about data privacy and its access restrictions due to existing licensing agreements, it is critical for spatio-temporal modeling to utilize decentralized data, yet leveraging the underlying relationships for improved performance. 

Although recent works in federated learning (FL)~\citep{kairouz2019advances} provides a solution for training a model with decentralized data on multiple devices, these works either do not consider the inherent spatio-temporal  dependencies~\citep{mcmahan2017communication,li2018federated,karimireddy2019scaffold} or only model it implicitly by imposing the graph structure in the regularization on model weights~\citep{smith2017federated}, the latter of which suffers from the limitation of regularization based methods due to the assumption that graphs only encode similarity of nodes~\citep{kipf2017semi}, and cannot operate in settings where only a fraction of devices are observed during training (\emph{inductive learning setting}). As a result, there is a need for an architecture for spatio-temporal data modeling which enables reliable computation on the edge, while maintaining the data decentralized. 

%v1
%Following the definition of cross-silo and cross-device federated learning in~\citep{kairouz2019advances}, we here define the \emph{cross-node} federated learning requirement of spatio-temporal data as the constraint that data in a network of nodes is generated locally on each node and remains decentralized. To bridge the gap between the modeling of spatio-temporal dynamics and the requirement of cross-node federated learning, we propose the \modelfullname~(\modelshortname), which effectively models the complex spatio-temporal dependencies under the cross-node federated learning constraint. \modelshortname~decomposes the modeling of temporal dependencies and spatial dependencies, where an encoder-decoder model on each device extracts the temporal features with local data, and a Graph Neural Network (GNN) based model on the server captures spatial dependencies among devices. 

To this end, leveraging recent works on federated learning \citep{kairouz2019advances}, we introduce the \emph{cross-node} federated learning requirement to ensure that data generated locally at a node remains decentralized. Specifically, our architecture -- \modelfullname~(\modelshortname), aims to effectively model the complex spatio-temporal dependencies under the cross-node federated learning constraint. For this, \modelshortname~decomposes the modeling of temporal and spatial dependencies using an encoder-decoder model on each device to extract the temporal features with local data, and a Graph Neural Network (GNN) based model on the server to capture spatial dependencies among devices.

% \begin{wrapfigure}{r}{0.5\textwidth}
% \vspace{-1em}
%   \begin{center}
%     \includegraphics[width=0.7\linewidth]{example-image-a}
%   \end{center}
%   \vspace{-1em}
%   \caption{\modelshortname.}
% \vspace{-1em}
% \end{wrapfigure}

%v1
%Instead of increasing the model complexity on devices, \modelshortname~enhances the prediction performance of on-device models with embeddings of spatial dependencies generated on the server-side without increasing the computation cost of edge devices. \modelshortname~utilizes FederatedAveraging~\citep{mcmahan2017communication} to train a shared temporal feature extractor over all devices, and split learning~\citep{singh2019detailed} for the training of the server-side GNN. Inspired by the success of alternating optimization in federated learning for learning global representations~\citep{liang2020think} and knowledge distillation~\citep{he2020group}, we alternate the training of on-device models and that of the server-side model to incorporate the spatial dependencies among nodes while reducing the communication cost by half approximately compared to the split learning method.

As compared to existing federated learning techniques that rely on regularization to incorporate spatial relationships, \modelshortname~ leverages an explicit graph structure using a graph neural network-based (GNNs) architecture, which leads to performance gains. However, the federated learning (data sharing) constraint means that the GNN cannot be trained in a centralized manner, since each node can only access the data stored on itself. To address this, \modelshortname~ employs Split Learning ~\citep{singh2019detailed} to train the spatial and temporal modules.
% , which also allows the flexibility for the node-side models to be heterogeneous (at the expense of computation cost) depending upon their local requirements. 
Further, to alleviate the associated high communication cost incurred by Split Learning, we propose an alternating optimization-based training procedure of these modules, which incurs only half the communication overhead as compared to a comparable Split Learning architecture. Here, we also use Federated Averaging (FedAvg)~\citep{mcmahan2017communication} to train a shared temporal feature extractor for all nodes, which leads to improved empirical performance.

Our main contributions are as follows :
\begin{enumerate}
   % \item We propose the \modelfullname~(\modelshortname), which can capture the complex spatio-temporal relationships among multiple nodes and to transfer to unobserved nodes in training while satisfying the requirement of cross-node federated learning.
    \item We propose \modelfullname~(\modelshortname), a GNN-based federated learning architecture that captures complex spatio-temporal relationships among multiple nodes while ensuring that the data generated locally remains decentralized at no extra computation cost at the edge devices.
    % \item \modelshortname~disentangles the modeling of local temporal dynamics on edge devices and the modeling of spatial dynamics on the central server, which enables clients to incorporate node relation information with no extra computation cost.
    \item Our modeling and training procedure enables GNN-based architectures to be used in federated learning settings. We achieve this by disentangling the modeling of local temporal dynamics on edge devices and spatial dynamics on the central server, and leverage an alternating optimization-based procedure for updating the spatial and temporal modules using Split Learning and Federated Averaging to enable effective GNN-based federated learning.
    
    %\item We evaluate the performance of \modelshortname~on a traffic flow prediction task as an example of spatio-temporal data modeling. Compared to baselines increasing the complexity of models on nodes, \modelshortname~achieves the largest improvement of prediction performance with no extra computation cost on edge devices and intermediate extra communication cost, while also showing the best performance in inductive learning.
    \item We demonstrate that \modelshortname~achieves the best prediction performance (both in transductive and inductive settings) at no extra computation cost on edge devices with modest communication cost, as compared to the related techniques on a traffic flow prediction task. 
    
\end{enumerate}
\section{Related Works}
Our method derives elements from graph neural networks, federated learning and privacy-preserving graph learning, we now discuss related works in these areas in relation to our work. 
\paragraph{Graph Neural Networks (GNNs).} GNNs have shown their superior performance on various learning tasks with graph-structured data, including graph embedding~\citep{hamilton2017inductive}, node classification~\citep{kipf2017semi}, spatio-temporal data modeling~\citep{yan2018spatial, li2018dcrnn_traffic, yu2018spatio} and multi-agent trajectory prediction~\citep{battaglia2016interaction, kipf2018neural, li2020generative}.
\update{
Recent GNN models~\citep{hamilton2017inductive, ying2018graph, you2019position, huang2018adaptive} also have sampling strategies and are able to scale on large graphs.
} While GNNs enjoy the benefit from strong inductive bias~\citep{battaglia2018relational, xu2019can}, most works require centralized data during the training and the inference processes.
\paragraph{Federated Learning (FL).} Federated learning is a machine learning setting where multiple clients train a model in collaboration with decentralized training data~\citep{kairouz2019advances}. It requires that the raw data of each client is stored locally without any exchange or transfer. However, the decentralized training data comes at the cost of less utilization due to the heterogeneous distributions of data on clients and the lack of information exchange among clients. Various optimization algorithms have been developed for federated learning on non-IID and unbalanced data \citep{mcmahan2017communication,li2018federated,karimireddy2019scaffold}. \cite{smith2017federated} propose a multi-task learning framework that captures relationships amongst data. While the above works mitigate the caveat of missing neighbors' information to some extent, they are not as effective as GNN models and still suffer from the absence of feature exchange and aggregation.

\paragraph{\update{Alternating Optimization.}}
\update{
Alternating optimization is a popular choice in non-convex optimization~\citep{agarwal2014learning, arora2014new, arora2015simple, MAL-058}. In the context of Federated Learning, \cite{liang2020think}~uses alternating optimization for learning a simple global model and reduces the number of communicated parameters, and \cite{he2020group}~uses alternating optimization for knowledge distillation from server models to edge models. In our work, we utilize alternating optimization to effectively train on-device modules and the server module jointly, which captures temporal and spatial relationships respectively.
}

\paragraph{Privacy-Preserving Graph Learning.} \cite{suzumura2019towards} and \cite{mei2019sgnn} use statistics of graph structures instead of node information exchange and aggregation to avoid the leakage of node information. Recent works have also incorporated graph learning models with privacy-preserving techniques such as Differential Privacy (DP), Secure Multi-Party Computation (MPC) and Homomorphic Encryption (HE).~\cite{zhou2020privacy} utilize MPC and HE when learning a GNN model for node classification with vertically split data to preserve silo-level privacy instead of node-level privacy. \cite{sajadmanesh2020differential} preprocesses the input raw data with DP before feeding it into a GNN model. Composing privacy-preserving techniques for graph learning can help build federated learning systems following the privacy-in-depth principle, wherein the privacy properties degrade as gracefully as possible if one technique fails~\citep{kairouz2019advances}.
\section{\modelfullname}
\subsection{Problem Formulation}
Given a dataset with a graph $\mathcal{G}=(\mathcal{V}, \mathcal{E})$, a feature tensor $\tX\in\mathbb{R}^{|\mathcal{V}|\times\dots}$ and a label tensor $\tY\in\mathbb{R}^{|\mathcal{V}|\times\dots}$, the task is defined on the dataset with $\tX$ as the input and $\tY$ as the prediction target. We consider learning a model under the cross-node federated learning constraint: node feature $\vx_i=\tX_{i,\dots}$, node label $\vy_i=\tY_{i,\dots}$, and model output $\hat{\vy}_i$ are only visible to the node $i$.

% One typical task that requires the node-level privacy is user-level trajectory prediction. In such a scenario, $\mathcal{V}$ is the set of users and $\mathcal{E}$ describes relations among users (e.g. $e_{ij}\in\mathcal{E}$ if and only if user $v_i$ and $v_j$ are friends). The feature tensor $\tX\in\mathbb{R}^{|\mathcal{V}|\times m\times D}$ gives all users' locations in the $D$-dim space during the past $m$ time steps, and the label $\tY\in\mathbb{R}^{|\mathcal{V}|\times n\times D}$ gives all users' locations in the future $n$ time steps. Since users' trajectory data is sensitive and users may not want to share it, it is necessary to consider the node-level privacy when designing the model.

One typical task that requires the cross-node federated learning constraint is the prediction of spatio-temporal data generated by a network of sensors. In such a scenario, $\mathcal{V}$ is the set of sensors and $\mathcal{E}$ describes relations among sensors (e.g. $e_{ij}\in \mathcal{E}$ if and only if the distance between $v_i$ and $v_j$ is below some threshold). The feature tensor $\vx_{i}\in\mathbb{R}^{m\times D}$ represents the $i$-th sensor's records in the $D$-dim space during the past $m$ time steps, and the label $\vy_{i}\in\mathbb{R}^{n\times D}$ represents the $i$-th sensor's records in the future $n$ time steps. Since records collected on different sensors owned by different users/organizations may not be allowed to be shared due to the need for edge computation or licensing issues on data access, it is necessary to design an algorithm modeling the spatio-temporal relation without any direct exchange of node-level data.

\subsection{Proposed Method}
% \begin{figure}[h]
%     \centering
%     \includegraphics[width=\linewidth]{figures/CNGNN.pdf}
%     \caption{The architecture of \modelshortname. \todo{Update all symbols and decompose it to 2 subfigures for macro/micro levels}}
%     \label{fig:modelarch}
% \end{figure}

\begin{figure}[t]
    \centering
    \begin{subfigure}[b]{0.5\linewidth}
         \centering
         \includegraphics[width=\textwidth]{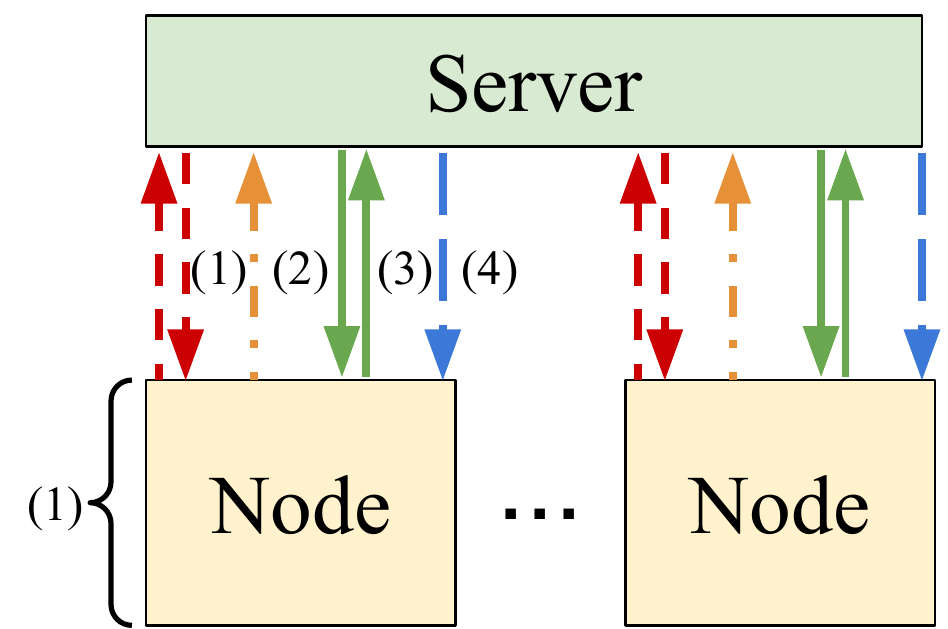}
         \caption{Overview of the training procedure.}
         \label{fig:alt-train}
     \end{subfigure}
     \hfill
     \begin{subfigure}[b]{\linewidth}
         \centering
         \includegraphics[width=\textwidth]{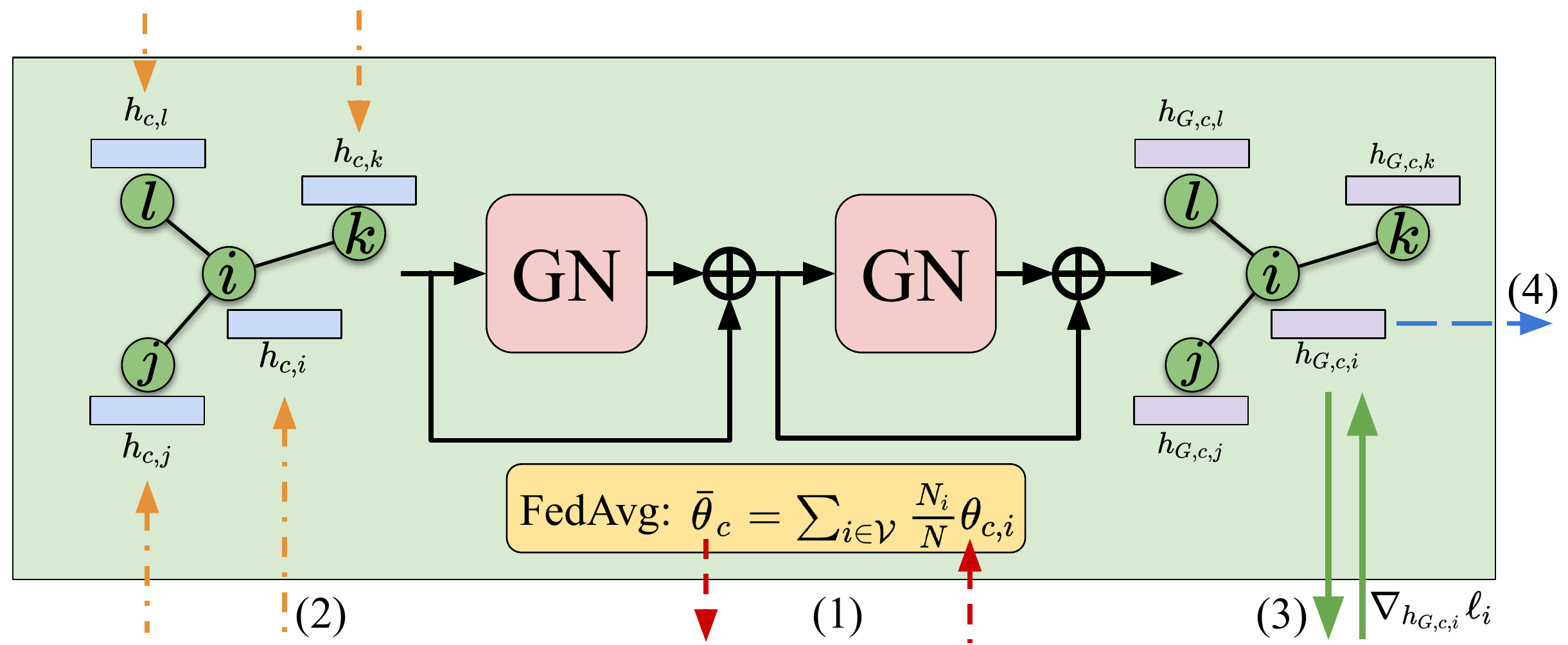}
         \caption{Server-side Graph Network (GN).}
         \label{fig:server-side-gn}
     \end{subfigure}
     \hfill
     \begin{subfigure}[b]{\linewidth}
         \centering
         \includegraphics[width=\textwidth]{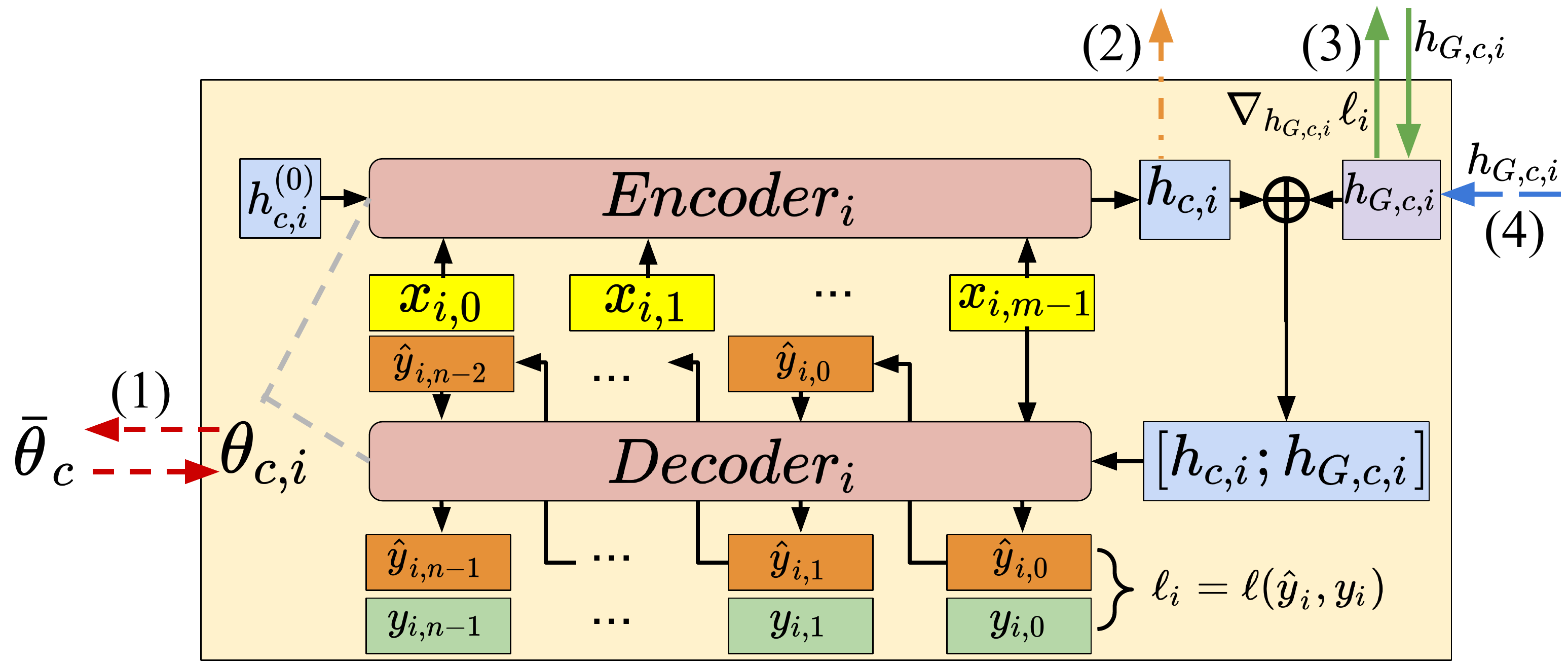}
         \caption{Encoder-decoder on the $i$-th node.}
         \label{fig:enc-dec-node}
     \end{subfigure}
    \caption{\modelfullname. (a) In each round of training, we alternately train models on nodes and the model on the server. More specifically, we sequentially execute: (1) Federated learning of on-node models. (2) Temporal encoding update. (3) Split Learning of GN. (4) On-node graph embedding update. (b) Detailed view of the server-side GN model for modeling spatial dependencies in data. (c) Detailed view of the encoder-decoder model on the $i$-th node.}
    \label{fig:model}
\end{figure}

\begin{algorithm*}[h]
	\caption{Training algorithm of \modelshortname~on the server side.}
	\label{alg:server-training}
	\begin{multicols}{2}
	\begin{flushleft}
	\textbf{Input:} Initial server-side GN weights $\vtheta^{(0)}_{GN}$, initial client model weights $\bar{\vtheta}^{(0)}_{c}=\{\bar{\vtheta}^{(0),enc}_{c}, \bar{\vtheta}^{(0), dec}_{c}\}$, the maximum number of global rounds $R_g$, the maximum number of client rounds $R_c$, the maximum number of server rounds $R_s$, server-side learning rate $\eta_s$, client learning rate $\eta_c$.\\
	\textbf{Output:} Trained server-side GN weights $\vtheta^{(R_g)}_{GN}$, trained client model weights $\bar{\vtheta}^{(R_g)}_{c}$.\\
	\textbf{Server executes:}
	\end{flushleft}
	\begin{algorithmic}[1]
% 		\For {$iteration=1,2,\ldots$}
% 			\For {$actor=1,2,\ldots,N$}
% 				\State Run policy $\pi_{\theta_{old}}$ in environment for $T$ time steps
% 				\State Compute advantage estimates $\hat{A}_{1},\ldots,\hat{A}_{T}$
% 			\EndFor
% 			\State Optimize surrogate $L$ wrt. $\theta$, with $K$ epochs and minibatch size $M\leq NT$
% 			\State $\theta_{old}\leftarrow\theta$
% 		\EndFor
        \State Initialize server-side GN weights with $\vtheta^{(0)}_{GN}$. Initialize client model weights with $\bar{\vtheta}^{(0)}_{c}$.
        \For {each node $i\in\mathcal{V}$ in parallel}
            \State Initialize client model $\vtheta^{(0)}_{c,i}=\bar{\vtheta}^{(0)}_{c}$.
            \State Initialize graph encoding on node $\vh_{G,c,i} = \vh^{(0)}_{G,c,i}$.
        \EndFor
        \For {global round $r_g=1,2,\ldots,R_g$}
            \State // \textbf{(1) Federated learning of on-node models.}
            \For {each client $i\in\mathcal{V}$ in parallel}
                \State $\vtheta_{c,i}$ $\leftarrow$ ClientUpdate($i$, $R_c$, $\eta_c$).
            \EndFor
            \State $\bar{\vtheta}_{c}^{(r_g)}\leftarrow\sum_{i\in\mathcal{V}}\frac{N_i}{N}\vtheta_{c,i}$.
            \For {each client $i\in\mathcal{V}$ in parallel}
                \State Initialize client model: $\vtheta^{(0)}_{c,i}=\bar{\vtheta}_{c}^{(r_g)}$.
            \EndFor
            \State // \textbf{(2) Temporal encoding update.}
            \For {each client $i\in\mathcal{V}$ in parallel}
                \State $\vh_{c,i}\leftarrow$ ClientEncode($i$).
            \EndFor
            \State // \textbf{(3) Split Learning of GN.}
            \State Initialize $\vtheta^{(r_g, 0)}_{GN} = \vtheta^{(r_g-1)}_{GN}$.
            \For {server round $r_s=1,2,\ldots,R_s$}
                \State $\{\vh_{G,c,i}|i\in\mathcal{V}\} \gets GN(\{\vh_{c,i}|i\in\mathcal{V}\};\vtheta^{(r_g, r_s-1)}_{GN})$.
                % \State $\nabla_{\vtheta^{(r_g, r_s-1)}_{GN}}\ell~\gets 0$.
                \For {each client $i\in\mathcal{V}$ in parallel}
                    \State \begin{varwidth}[t]{\linewidth}
                    $\nabla_{\vh_{G,c,i}}\ell_{i}\leftarrow$ ClientBackward(\par
                    \hskip\algorithmicindent $i$,$\vh_{G,c,i}$).
                    \end{varwidth}
                    \State \begin{varwidth}[t]{\linewidth}
                    $\nabla_{\vtheta^{(r_g, r_s-1)}_{GN}}\ell_i~\gets\vh_{G,c,i}$.backward(\par
                    \hskip\algorithmicindent$\nabla_{h_{G,c,i}}\ell_{i}$).
                    \end{varwidth}
                    % \State \begin{varwidth}[t]{\linewidth}
                    % $\nabla_{\vtheta^{(r_g, r_s-1)}_{GN}}\ell~\gets \nabla_{\vtheta^{(r_g, r_s-1)}_{GN}}\ell$\par
                    % \hskip\algorithmicindent $+\nabla_{\vtheta^{(r_g, r_s-1)}_{GN}}\ell_i$.
                    % \end{varwidth}
                \EndFor
                \State \begin{varwidth}[t]{\linewidth}
                $\nabla_{\vtheta^{(r_g, r_s-1)}_{GN}}\ell~\gets \sum_{i\in\mathcal{V}}\nabla_{\vtheta^{(r_g, r_s-1)}_{GN}}\ell_i$.
                \end{varwidth}
                \State \begin{varwidth}[t]{\linewidth} $\vtheta^{(r_g, r_s)}_{GN}\leftarrow\vtheta^{(r_g, r_s-1)}_{GN}$\par
                \hskip\algorithmicindent - $\eta_{s}\nabla_{\vtheta^{(r_g, r_s-1)}_{GN}}\ell$.
                \end{varwidth}
            \EndFor
            \State $\vtheta^{(r_g)}_{GN}\leftarrow\vtheta^{(r_g, R_s)}_{GN}$.
            \State // \textbf{(4) On-node graph embedding update.}
            \State \begin{varwidth}[t]{\linewidth}
            $\{\vh_{G,c,i}|i\in\mathcal{V}\} \gets$\par
            \hskip\algorithmicindent $GN(\{\vh_{c,i}|i\in\mathcal{V}\};\vtheta^{(r_g)}_{GN})$.
            \end{varwidth}
            \For {each client $i\in\mathcal{V}$ in parallel}
                \State Set graph encoding on client as $\vh_{G,c,i}$.
            \EndFor
        \EndFor
	\end{algorithmic}
	\end{multicols}
\end{algorithm*}

\begin{algorithm*}[h]
    \caption{Training algorithm of \modelshortname~on the client side.}
    \label{alg:client-training}
	\begin{multicols}{2}
	\begin{flushleft}
	\textbf{ClientUpdate($i$, $R_c$, $\eta_c$):}
	\end{flushleft}
	\begin{algorithmic}[1]
	    \For {client round $r_c=1,2,\ldots,R_c$}
	        \State $\vh^{(r_c)}_{c,i}\leftarrow Encoder_i(\vx_{i};\vtheta^{(r_c-1),enc}_{c,i})$.
	        \State \begin{varwidth}[t]{\linewidth}
	        $\hat{\vy}_i\leftarrow Decoder_i($\par
	        \hskip\algorithmicindent $x_{i,m},[\vh^{(r_c)}_{c,i};\vh_{G,c,i}];\vtheta^{(r_c-1),dec}_{c,i})$.
	        \end{varwidth}
	        \State $\ell_i\leftarrow\ell(\hat{\vy}_i,\vy)$.
	        \State $\vtheta^{(r_c)}_{c,i}\leftarrow\vtheta^{(r_c-1)}_{c,i} - \eta_{c}\nabla_{\vtheta^{(r_c-1)}_{c,i}}\ell_i$.
	    \EndFor
	    \State $\vtheta_{c,i}=\vtheta^{(R_c)}_{c,i}$.
	    \State \Return $\vtheta_{c,i}$ to server.
	\end{algorithmic}
	\begin{flushleft}
	\textbf{ClientEncode($i$):}
	\end{flushleft}
	\begin{algorithmic}[1]
	    \State \Return $\vh_{c,i} = Encoder_i(\vx_{i};\vtheta^{enc}_{c,i})$ to server.
	\end{algorithmic}
	\begin{flushleft}
	\textbf{ClientBackward($i,h_{G,c,i}$):}
    \end{flushleft}
	\begin{algorithmic}[1]
	    \State $\hat{\vy}_i\leftarrow Decoder_i(x_{i,m},[h_{c,i};h_{G,c,i}];\vtheta^{dec}_{c,i})$.
	   \State $\ell_i\leftarrow\ell(\hat{\vy}_i,\vy)$.
	   \State \Return $\nabla_{\vh_{G,c,i}}\ell_i$ to server.
	\end{algorithmic}
	\end{multicols}
\end{algorithm*}

We now introduce our proposed \modelfullname~(\modelshortname) model. Here, we begin by disentangling the modeling of node-level temporal dynamics and server-level spatial dynamics as follows: (i) (\fref{fig:enc-dec-node}) on each node, an encoder-decoder model extracts temporal features from data on the node and makes predictions; (ii) (\fref{fig:server-side-gn}) on the central server, a Graph Network (GN)~\citep{battaglia2018relational} propagates extracted node temporal features and outputs node embeddings, which incorporate the relationship information amongst nodes. (i) has access to the not shareable node data and is executed on each node locally. (ii) only involves the upload and download of smashed features and gradients instead of the raw data on nodes. This decomposition enables the exchange and aggregation of node information under the cross-node federated learning constraint.

\subsubsection{Modeling of Node-Level Temporal Dynamics}
We modify the Gated Recurrent Unit (GRU) based encoder-decoder architecture in \citep{cho2014learning} for the modeling of node-level temporal dynamics on each node. Given an input sequence $\vx_i\in\mathbb{R}^{m\times D}$ on the $i$-th node, an encoder sequentially reads the whole sequence and outputs the hidden state $\vh_{c,i}$ as the summary of the input sequence according to \eref{eq:encoder}.
\begin{equation}
    \begin{aligned}
    \vh_{c,i} &= Encoder_{i}(\vx_{i}, \vh^{(0)}_{c, i}),
    \end{aligned}
    \label{eq:encoder}
\end{equation} where $\vh^{(0)}_{c,i}$ is a zero-valued initial hidden state vector.

To incorporate the spatial dynamics into the prediction model of each node, we concatenate $\vh_{c,i}$ with the node embedding $\vh_{G,c,i}$ generated from the procedure described in \ref{subsubsec:modeling-of-global}, which contains spatial information, as the initial state vector of the decoder. The decoder generates the prediction $\hat{\vy}_i$ in an auto-regressive way starting from the last frame of the input sequence $x_{i,m}$ with the concatenated hidden state vector.
\begin{equation}
    \hat{\vy}_i = Decoder_i(x_{i,m}, [\vh_{c,i}; \vh_{G,c,i}]).
\end{equation} We choose the mean squared error (MSE) between the prediction and the ground truth values as the loss function, which is evaluated on each node locally.

\subsubsection{Modeling of Spatial Dynamics}
\label{subsubsec:modeling-of-global}
To capture the complex spatial dynamics, we adopt Graph Networks (GNs) proposed in \citep{battaglia2018relational} to generate node embeddings containing the relational information of all nodes. The central server collects the hidden state from all nodes $\{\vh_{c,i}~\vert~i\in\mathcal{V}\}$ as the input to the GN. Each layer of GN updates the input features as follows:
\begin{equation}
    \begin{array}{ll}
\mathbf{e}_{k}^{\prime}=\phi^{e}\left(\mathbf{e}_{k}, \mathbf{v}_{r_{k}}, \mathbf{v}_{s_{k}}, \mathbf{u}\right) & \overline{\mathbf{e}}_{i}^{\prime}=\rho^{e \rightarrow v}\left(E_{i}^{\prime}\right) \\
\mathbf{v}_{i}^{\prime}=\phi^{v}\left(\overline{\mathbf{e}}_{i}^{\prime}, \mathbf{v}_{i}, \mathbf{u}\right) & \overline{\mathbf{e}}^{\prime}=\rho^{e \rightarrow u}\left(E^{\prime}\right) \\
\mathbf{u}^{\prime}=\phi^{u}\left(\overline{\mathbf{e}}^{\prime}, \overline{\mathbf{v}}^{\prime}, \mathbf{u}\right) & \overline{\mathbf{v}}^{\prime}=\rho^{v \rightarrow u}\left(V^{\prime}\right)
\end{array},
\end{equation}
where $\mathbf{e}_{k},\mathbf{v}_i,\mathbf{u}$ are edge features, node features and global features respectively. $\phi^e, \phi^v, \phi^u$ are neural networks. $\rho^{e \rightarrow v}, \rho^{e \rightarrow u}, \rho^{v \rightarrow u}$ are aggregation functions such as summation. As shown in \fref{fig:server-side-gn}, we choose a 2-layer GN with residual connections for all experiments. \update{We set $\mathbf{v}_i=\vh_{c,i}$, $\mathbf{e}_{k}=W_{r_k,s_k}$ ($W$ is the adjacency matrix) , and assign the empty vector to $\mathbf{u}$ as the input of the first GN layer.} The server-side GN outputs embeddings $\{\vh_{G,c,i}~\vert~i\in\mathcal{V}\}$ for all nodes, and sends the embedding of each node correspondingly. 

\subsubsection{Alternating Training of Node-Level and Spatial Models}
\label{subsubsec:alternating-training}
One challenge brought about by the cross-node federated learning requirement and the server-side GN model is the high communication cost in the training stage. Since we distribute different parts of the model on different devices, Split Learning proposed by \citep{singh2019detailed} is a potential solution for training, where hidden vectors and gradients are communicated among devices. However, when we simply train the model end-to-end via Split Learning, the central server needs to receive hidden states from all nodes and to send node embeddings to all nodes in the forward propagation, then it must receive gradients of node embeddings from all nodes and send back gradients of hidden states to all nodes in the backward propagation. Assume all hidden states and node embeddings have the same size $S$, the total amount of data transmitted in each training round of the GN model is $4|\mathcal{V}|S$.

To alleviate the high communication cost in the training stage, we instead alternately train models on nodes and the GN model on the server. More specifically, in each round of training, we (1) fix the node embedding $\vh_{G,c,i}$ and optimize the encoder-decoder model for $R_c$ rounds, then (2) we optimize the GN model while fixing all models on nodes. Since models on nodes are fixed, $\vh_{c,i}$ stays constant during the training of the GN model, and the server only needs to fetch $\vh_{c,i}$ from nodes before the training of GN starts and only to communicate node embeddings and gradients. Therefore, the average amount of data transmitted in each round for $Rs$ rounds of training of the GN model reduces to $\frac{2 + 2R_s}{R_s}|\mathcal{V}|S$. We provide more details of the training procedure in \aref{alg:server-training} and \aref{alg:client-training}.

To more effectively extract temporal features from each node, we also train the encoder-decoder models on nodes with the FedAvg algorithm proposed in \citep{mcmahan2017communication}. This enables all nodes to share the same feature extractor and thus share a joint hidden space of temporal features, which avoids the potential overfitting of models on nodes and demonstrates faster convergence and better prediction performance empirically.

\section{Experiments}

We evaluate the performance of \modelshortname~and all baseline methods on the traffic forecasting task, which is an important application for spatio-temporal data modeling.
% The aim of our paper is to enable Graph Neural Network(GNN)-based architecture, which has shown its success in modeling centralized spatio-temporal data, under the setting of FL. 
The primary challenge in FL is to respect constraints on data sharing and manipulation. These constraints can occur in scenarios where data contains sensitive information, such as financial data owned by different institutions. Due to the sensitivity of data, datasets from such scenarios are proprietary and hardly offer public access. Therefore, we demonstrate the applicability of our proposed model on the traffic dataset, which is a good representative example of data with spatio-temporal correlations, and has been extensively studied in spatio-temporal forecasting works without FL constraints~\citep{li2018dcrnn_traffic,yu2018spatio}. Our proposed model is general and applicable to other spatio-temporal datasets with sensitive information.

We reuse the following two real-world large-scale datasets in~\citep{li2018dcrnn_traffic} and follow the same preprocessing procedures: (1) \textbf{PEMS-BAY}: This dataset contains the traffic speed readings from 325 sensors in the Bay Area over 6 months from Jan 1st, 2017 to May 31st, 2017. (2) \textbf{METR-LA}: This dataset contains the traffic speed readings from 207 loop detectors installed on the highway of Los Angeles County over 4 months from Mar 1st, 2012 to Jun 30th, 2012.
% \begin{enumerate}
%     \item \textbf{PEMS-BAY}: This dataset contains the traffic speed readings from 325 sensors in the Bay Area over 6 months from Jan 1st 2017 to May 31th 2017.
%     \item \textbf{METR-LA}: This dataset contains the traffic speed readings from 207 loop detectors installed on the highway of Los Angeles County over 4 months from Mar 1st 2012 to Jun 30th 2012.
% \end{enumerate}

For both datasets, we construct the adjacency matrix of sensors using the Gaussian kernel with a threshold: $W_{i,j}=d_{i,j}~\mathrm{if}~d_{i,j}>=\kappa~\mathrm{else}~0$, where $d_{i,j} = \exp{(-\frac{\mathrm{dist}(v_i,v_j)^2}{\sigma^2})}$, $\mathrm{dist}(v_i, v_j)$ is the road network distance from sensor $v_i$ to sensor $v_j$, $\sigma$ is the standard deviation of distances and $\kappa$ is the threshold. We set $\kappa=0.1$ for both datasets.

We aggregate traffic speed readings in both datasets into 5-minute windows and truncate the whole sequence to multiple sequences with length 24. The forecasting task is to predict the traffic speed in the following 12 steps of each sequence given the first 12 steps. We show the statistics of both datasets in \tref{tab:stat}.

\begin{table}[htbp]
\centering
\caption{Statistics of datasets PEMS-BAY and METR-LA.}
\label{tab:stat}
\resizebox{\linewidth}{!}{%
\begin{tabular}{@{}cccccc@{}}
\toprule
Dataset  & \# Nodes & \# Directed Edges & \# Train Seq & \# Val Seq & \# Test Seq \\ \midrule
PEMS-BAY & 325      & 2369              & 36465         & 5209       &  10419           \\
METR-LA  & 207      & 1515              & 23974         & 3425       &  6850           \\ \bottomrule
\end{tabular}
}
\end{table}
\vspace{-4pt}

\subsection{Spatio-Temporal Data Modeling: Traffic Flow Forecasting}

\paragraph{Baselines}
Here we introduce the settings of baselines and our proposed model \modelshortname. Unless noted otherwise, all models are optimized using the Adam optimizer with the learning rate 1e-3.
\begin{itemize}
    \item \textbf{GRU (centralized)}: Gated Recurrent Unit (GRU) model trained with centralized sensor data. The GRU model with 63K parameters is a 1-layer GRU with hidden dimension 100, and the GRU model with 727K parameters is a 2-layer GRU with hidden dimension 200.
    \item \textbf{GRU + GN (centralized)}: a model directly combining GRU and GN trained with centralized data, whose architecture is similar to \modelshortname~but all GRU modules on nodes always share the same weights. We see its performance as the upper bound of the performance of \modelshortname.
    \item  \textbf{GRU (local)}: for each node we train a GRU model with only the local data on it.
    \item \textbf{GRU + FedAvg}: a GRU model trained with the Federated Averaging algorithm~\citep{mcmahan2017communication}. We select 1 as the number of local epochs.
    \item \textbf{GRU + FMTL}: for each node we train a GRU model using the federated multi-task learning (FMTL) with cluster regularization~\citep{smith2017federated} given by the adjacency matrix. More specifically, the cluster regularization (without the L2-norm regularization term) takes the following form:
\begin{equation}
    \mathcal{R}(\mW,\mathbf{\Omega}) = \lambda\mathrm{tr}(\mW\mathbf{\Omega}\mW^T).
    \label{eq:cluster}
\end{equation}
Given the constructed adjacency matrix $\mA$, $\mathbf{\Omega}=\frac{1}{|\mathcal{V}|}(\mD - \mA)=\frac{1}{|\mathcal{V}|}\mL$, where $\mD$ is the degree matrix and $\mL$ is the Laplacian matrix. Equation~\ref{eq:cluster} can be reformulated as:
\begin{equation}
    \begin{aligned}
    \mathcal{R}(\mW,\mathbf{\Omega}) &= \lambda\mathrm{tr}(\mW\mathbf{\Omega}\mW^T)\\
    &= \frac{\lambda}{|\mathcal{V}|}\mathrm{tr}(\mW \mL \mW^T) \\
    &= \frac{\lambda}{|\mathcal{V}|}\mathrm{tr}(\sum_{i\in\mathcal{V}}\vw_i\sum_{j\neq i}a_{ij}\vw^T_i - \sum_{j\neq i}\vw_i a_{ij}\vw^T_j)\\
    &= \lambda_1 (\sum_{i\in\mathcal{V}}\sum_{j\neq i}\alpha_{i,j}\langle \vw_i, \vw_i - \vw_j \rangle).
    \end{aligned}
    \label{eq:cluster-reg}
\end{equation} We implement the cluster regularization via sharing model weights between each pair of nodes connected by an edge and select $\lambda_1=0.1$.
For each baseline, we have 2 variants of the GRU model to show the effect of on-device model complexity: one with 63K parameters and the other with 727K parameters. For \modelshortname, the encoder-decoder model on each node has 64K parameters and the GN model has 1M parameters.

\item \textbf{\modelshortname} We use a GRU-based encoder-decoder model as the model on nodes, which has 1 GRU layer and hidden dimension 64. We use a 2-layer Graph Network (GN) with residual connections as the Graph Neural Network model on the server side. We use the same network architecture for the edge/node/global update function in each GN layer: a multi-layer perceptron (MLP) with 3 hidden layers, whose sizes are [256, 256, 128] respectively. We choose $R_c=1, R_s=20$ for experiments on PEMS-BAY, and $R_c=1, R_s=1$ for METR-LA.

\end{itemize}

\paragraph{Calculation of Communication Cost}

We denote $R$ as the number of communication rounds for one model to reach the lowest validation error in the training stage.

\paragraph{GRU + FMTL} Using Equation \ref{eq:cluster-reg}, in each communication round, each pair of nodes exchange their model weights, thus the total communicated data amount is calculated as:
\begin{equation}
    R\times \mathrm{\# nonself~directed~edges} \times \mathrm{size~of~node~model~weights}.
\end{equation}
We list corresponding parameters in \tref{tab:params-comm-cost}.

\paragraph{\update{\modelshortname~(AT + FedAvg)}} In each communication round, the central server fetches and sends back model weights to each node for Federated Averaging, and transmits hidden vectors and gradients for Split Learning. The total communicated data amount is calculated as:
\begin{equation}
\begin{aligned}
    R&\times (\mathrm{\#nodes} \times\mathrm{size~of~node~model~weights}\times 2\\
    &+ (1 + 2 * \mathrm{server~round} + 1)\times \mathrm{\#nodes}\times \mathrm{hidden~state~size}).
\end{aligned}
\end{equation}
We list corresponding parameters in \tref{tab:params-comm-cost-ours}.

\paragraph{\update{\modelshortname~(SL)}}
\update{
In each communication round, each node sends and fetches hidden vectors and graidents twice (one for encoder, the other for decoder) and the total communicated data amount is:
\begin{equation}
    R \times 2 \times 2 \times \mathrm{\#nodes}\times \mathrm{hidden~state~size}.
\end{equation}
We list corresponding parameters in \tref{tab:params-comm-cost-ours-sl}.
}

\paragraph{\update{\modelshortname~(SL + FedAvg)}}
\update{
Compared to \modelshortname~(SL), the method has extra communcation cost for FedAvg in each round, thus the total communicated data amount is:
\begin{equation}
    \begin{aligned}
    R&\times (\mathrm{\#nodes} \times\mathrm{size~of~node~model~weights} \times 2\\
    &+ 2 \times 2\times \mathrm{\#nodes}\times \mathrm{hidden~state~size}).
\end{aligned}
\end{equation}
We list corresponding parameters in \tref{tab:params-comm-cost-ours-sl-fedavg}.
}

\paragraph{\update{\modelshortname~(AT, w/o FedAvg)}}
\update{
Compared to \modelshortname~(AT + FedAvg), there is no communcation cost for the FedAvg part, thus the total communcated data amount is:
\begin{equation}
    \begin{aligned}
    R&\times (1 + 2 * \mathrm{server~round} + 1)\times \mathrm{\#nodes}\times \mathrm{hidden~state~size}.
\end{aligned}
\end{equation}
We list corresponding parameters in \tref{tab:params-comm-cost-ours-at-wo-fedavg}.
}

% \tref{tab:params-comm-cost}, \tref{tab:params-comm-cost-ours}, \tref{tab:params-comm-cost-ours-sl}, \tref{tab:params-comm-cost-ours-sl-fedavg} and \tref{tab:params-comm-cost-ours-at-wo-fedavg} demonstrate all parameters used in calculating the communication cost of GRU + FMTL and \update{\modelshortname~(AT + FedAvg)}.

\begin{table}[htbp]
% \tiny
\centering
\caption{Parameters used for calculating the communication cost of GRU + FMTL.}
\label{tab:params-comm-cost}
% \begin{tabular}{@{}cccccccc@{}}
% \toprule
% \multirow{2}{*}{Method} & \multirow{2}{*}{\begin{tabular}[c]{@{}c@{}}Node Model\\Weights Size (GB)\end{tabular}} & \multicolumn{3}{c}{PEMS-BAY}                                           & \multicolumn{3}{c}{METR-LA}                                             \\ \cmidrule(l){3-8} 
%                         &                                                                                          & \begin{tabular}[c]{@{}c@{}}\# Nonself \\Directed Edges\end{tabular}  & $R$ & \begin{tabular}[c]{@{}c@{}}Train Comm\\ Cost (GB)\end{tabular} & \begin{tabular}[c]{@{}c@{}}\# Nonself \\Directed Edges\end{tabular} & $R$   & \begin{tabular}[c]{@{}c@{}}Train Comm\\ Cost (GB)\end{tabular} \\ \midrule
% GRU (63K) + FMTL        & 2.347E-4                                                                                    & 2369 & 104 & 57.823                                                         & 1515 & 279 & 99.201                                                         \\
% GRU (727K) + FMTL       & 2.708E-3                                                                                    & 2369 & 56 & 359.292                                                        & 1515 & 176 & 722.137                                                        \\ \bottomrule
% \end{tabular}
\resizebox{\linewidth}{!}{%
\begin{tabular}{@{}cccc@{}}
\toprule
\multicolumn{2}{c}{Method} & GRU (63K) + FMTL & GRU (727K) + FMTL \\ \midrule
\multicolumn{2}{c}{Node Model Weights Size (GB)} & 2.347E-4 & 2.708E-3 \\ \midrule
\multirow{3}{*}{PEMS-BAY} & \#Nonself Directed Edges & \multicolumn{2}{c}{2369} \\
 & R & 104 & 56 \\
 & Train Comm Cost (GB) & 57.823 & 359.292 \\ \midrule
\multirow{3}{*}{METR-LA} & \#Nonself Directed Edges & \multicolumn{2}{c}{1515} \\
 & R & 279 & 176 \\
 & Train Comm Cost (GB) & 99.201 & 722.137 \\ \bottomrule
\end{tabular}
}
\end{table}

\begin{table}[htbp]
\centering
\caption{Parameters used for calculating the communication cost of \update{\modelshortname~(AT + FedAvg)}.}
\label{tab:params-comm-cost-ours}
\begin{tabular}{@{}ccc@{}}
\toprule
\begin{tabular}[c]{@{}c@{}}Node Model\\ Weights Size (GB)\end{tabular} & \multicolumn{2}{c}{2.384E-4} \\ \midrule
\multirow{5}{*}{PEMS-BAY} & \#Nodes & 325 \\
 & Hidden State Size (GB) & 2.173E-3 \\
 & Server Round & 20 \\
 & R & 2 \\
 & Train Comm Cost (GB) & 237.654 \\ \midrule
\multirow{5}{*}{METR-LA} & \#Nodes & 207 \\
 & Hidden State Size (GB) & 1.429E-3 \\
 & Server Round & 1 \\
 & R & 46 \\
 & Train Comm Cost (GB) & 222.246 \\ \bottomrule
\end{tabular}
\end{table}

\begin{table}[htbp]
\centering
\caption{Parameters used for calculating the communication cost of \update{\modelshortname~(SL)}.}
\label{tab:params-comm-cost-ours-sl}
\begin{tabular}{@{}ccc@{}}
\toprule
\multirow{4}{*}{PEMS-BAY} & \#Nodes & 325 \\
 & Hidden State Size (GB) & 2.173E-3 \\
 & R & 31 \\
 & Train Comm Cost (GB) & 350.366 \\ \midrule
\multirow{4}{*}{METR-LA} & \#Nodes & 207 \\
 & Hidden State Size (GB) & 1.429E-3 \\
 & R & 65 \\
 & Train Comm Cost (GB) & 307.627 \\ \bottomrule
\end{tabular}
\end{table}

\begin{table}[htbp]
\centering
\caption{Parameters used for calculating the communication cost of \update{\modelshortname~(SL + FedAvg)}.}
\label{tab:params-comm-cost-ours-sl-fedavg}
\begin{tabular}{@{}ccc@{}}
\toprule
\begin{tabular}[c]{@{}c@{}}Node Model\\ Weights Size (GB)\end{tabular} & \multicolumn{2}{c}{2.384E-4} \\ \midrule
\multirow{4}{*}{PEMS-BAY} & \#Nodes & 325 \\
 & Hidden State Size (GB) & 2.173E-3 \\
 & R & 7 \\
 & Train Comm Cost (GB) & 80.200 \\ \midrule
\multirow{4}{*}{METR-LA} & \#Nodes & 207 \\
 & Hidden State Size (GB) & 1.429E-3 \\
 & R & 71 \\
 & Train Comm Cost (GB) & 343.031 \\ \bottomrule
\end{tabular}
\end{table}

\begin{table}[htbp]
\centering
\caption{Parameters used for calculating the communication cost of \update{\modelshortname~(AT, w/o FedAvg)}.}
\label{tab:params-comm-cost-ours-at-wo-fedavg}
\begin{tabular}{@{}ccc@{}}
\toprule
\multirow{5}{*}{PEMS-BAY} & \#Nodes & 325 \\
 & Hidden State Size (GB) & 2.173E-3 \\
 & Server Round & 20 \\
 & R & 44 \\
 & Train Comm Cost (GB) & 5221.576 \\ \midrule
\multirow{5}{*}{METR-LA} & \#Nodes & 207 \\
 & Hidden State Size (GB) & 1.429E-3 \\
 & Server Round & 1 \\
 & R & 49 \\
 & Train Comm Cost (GB) & 2434.985 \\ \bottomrule
\end{tabular}
\end{table}

\begin{table}[htbp]
\centering
\vspace{-2pt}
\caption{Comparison of performance on the traffic flow forecasting task. We use the Rooted Mean Squared Error (\textbf{RMSE}) to evaluate the forecasting performance.}
\label{tab:traffic-forecasting}
\begin{tabular}{@{}ccc@{}}
\toprule
Method              & PEMS-BAY & METR-LA\\ \midrule
GRU (centralized, 63K) &  4.124 & 11.730\\
GRU (centralized, 727K) &  4.128 & 11.787\\ 
\update{\begin{tabular}{@{}c@{}}GRU + GN \\ (centralized, 64K + 1M)\end{tabular}} &  \update{\textbf{3.816}} & \update{\textbf{11.471}}\\
\cmidrule(l){1-3}
GRU (local, 63K) & 4.010 & 11.801\\
GRU (local, 727K) & 4.152 & 12.224\\
GRU (63K) + FedAvg  &  4.512& 12.132 \\
GRU (727K) + FedAvg &  4.432& 12.058\\
GRU (63K) + FMTL    &  3.961 & 11.548\\
GRU (727K) + FMTL   &  3.955 & 11.570 \\ \cmidrule(l){1-3}
\modelshortname~(64K + 1M)   & \textbf{3.822} & \textbf{11.487}            \\
\bottomrule
\end{tabular}
\end{table}

\begin{table}[htbp]
\centering
\caption{Comparison of the computation cost on edge devices and the communication cost. We use the amount of floating point operations (\textbf{FLOPS}) to measure the computational cost of models on edge devices. We also show the total size of data/parameters transmitted in the training stage (\textbf{Train Comm Cost}) until the model reaches its lowest validation error.}
\label{tab:comp-comm-cost}
\resizebox{\linewidth}{!}{%
\begin{tabular}{@{}cccccc@{}}
\toprule
\multirow{2}{*}{Method} & \multirow{2}{*}{\begin{tabular}[c]{@{}c@{}}Comp Cost On \\ Device (GFLOPS)\end{tabular}} & \multicolumn{2}{c}{PEMS-BAY}                                           & \multicolumn{2}{c}{METR-LA}                                             \\ \cmidrule(l){3-6} 
                        &                                                                                          & RMSE  & \begin{tabular}[c]{@{}c@{}}Train Comm\\ Cost (GB)\end{tabular} & RMSE   & \begin{tabular}[c]{@{}c@{}}Train Comm\\ Cost (GB)\end{tabular} \\ \midrule
GRU (63K) + FMTL        & 0.159                                                                                    & 3.961 & 57.823                                                         & 11.548 & 99.201                                                         \\
GRU (727K) + FMTL       & 1.821                                                                                    & 3.955 & 359.292                                                        & 11.570 & 722.137                                                        \\
CNFGNN (64K + 1M)       & 0.162                                                                                    & \textbf{3.822} & 237.654                                                        & \textbf{11.487} & 222.246\\ \bottomrule %647.280                                                        \\ \bottomrule
\end{tabular}
}
\end{table}
\vspace{-4pt}

\paragraph{Discussion} \tref{tab:traffic-forecasting} shows the comparison of forecasting performance and \tref{tab:comp-comm-cost} shows the comparison of computation cost on device and communication cost of \modelshortname~and baselines. We make the following observations. Firstly, when we compare the best forecasting performance of each baseline over the 2 GRU variants, GRU trained with FedAvg performs the worst in terms of forecasting performance compared to GRU trained with centralized data and GRU trained with local data (4.432 vs 4.010/4.124 on PEMS-BAY and 12.058 vs 11.730/11.801 on METR-LA), showing that the data distributions on different nodes are highly heterogeneous, and training one single model ignoring the heterogeneity is suboptimal. 

Secondly, both the GRU+FMTL baseline and \modelshortname~consider the spatial relations among nodes and show better forecasting performance than baselines without relation information. This shows that the modeling of spatial dependencies is critical for the forecasting task. 

Lastly, \modelshortname~achieves the lowest forecasting error on both datasets. The baselines that increases the complexity of on-device models (GRU (727K) + FMTL) gains slight or even no improvement at the cost of higher computation cost on edge devices and larger communication cost. However,  due to its effective modeling of spatial dependencies in data, \modelshortname~not only has the largest improvement of forecasting performance, but also keeps the computation cost on devices almost unchanged and maintains modest communication cost compared to baselines increasing the model complexity on devices.

\subsection{Inductive Learning on Unseen Nodes}
\label{subsec:inductive}
\begin{table*}[htbp]
\centering
\caption{Inductive learning performance measured with rooted mean squared error (RMSE).}
\label{tab:inductive}
% \resizebox{\linewidth}{!}{%
\begin{tabular}{@{}ccccccccccc@{}}
\toprule
\multirow{2}{*}{Method} & \multicolumn{5}{c}{PEMS-BAY} & \multicolumn{5}{c}{METR-LA} \\ \cmidrule(l){2-11} 
                    &5\%    & 25\%       & 50\%       & 75\% & 90\%  &5\%   & 25\%       & 50\%       & 75\%   &90\%    \\ \midrule
GRU (63K) + FedAvg    &\textbf{5.087} & 4.863 & 4.847 & 4.859&4.866& \textbf{12.128}&\textbf{11.993} & 12.104 &  12.014&12.016 \\
\modelshortname~(64K + 1M)    &5.869& \textbf{4.541} & \textbf{4.598} & \textbf{4.197}&\textbf{3.942}&13.931 & 12.013    & \textbf{11.815} & \textbf{11.676}&\textbf{11.629}\\ \bottomrule
\end{tabular}
% }
\end{table*}

\paragraph{Set-up} Another advantage of \modelshortname~is that it can conduct inductive learning and generalize to larger graphs with nodes unobserved during the training stage. We evaluate the performance of \modelshortname~under the following inductive learning setting: for each dataset, we first sort all sensors based on longitudes, then use the subgraph on the first $\eta\%$ of sensors to train the model and evaluate the trained model on the entire graph. For each dataset we select $\eta\%=25\%,50\%,75\%$. Over all baselines following the cross-node federated learning constraint, GRU (local) and GRU + FMTL requires training new models on unseen nodes and only GRU + FedAvg is applicable to the inductive learning setting.

\paragraph{Discussion} \tref{tab:inductive} shows the performance of inductive learning of \modelshortname~and GRU + FedAvg baseline on both datasets. We observe that under most settings, \modelshortname~outperforms the GRU + FedAvg baseline (except on the METR-LA dataset with 25\% nodes observed in training, where both models perform similarly), showing that \modelshortname~has the stronger ability of generalization.

% \subsection{\update{Inductive Learning}}
% \label{app:inductive}
\begin{figure}[t]
    \centering
    \begin{subfigure}[b]{\linewidth}
        \centering
        \includegraphics[width=\textwidth]{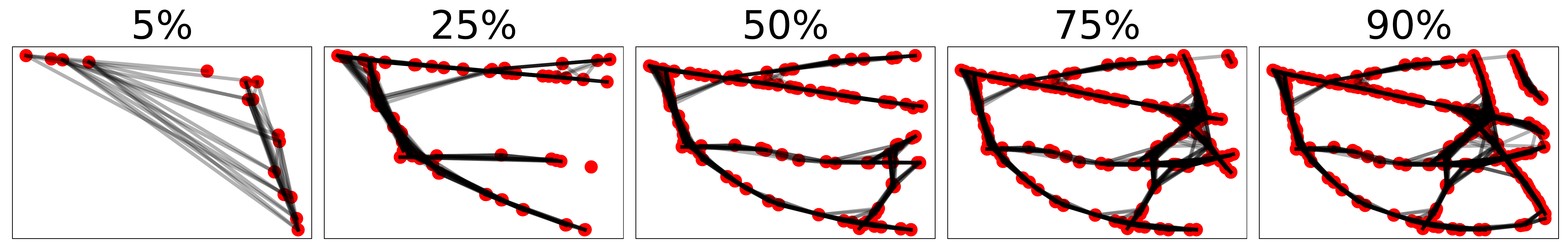}
        \caption{\update{PEMS-BAY}}
    \end{subfigure}
    \begin{subfigure}[b]{\linewidth}
        \centering
        \includegraphics[width=\textwidth]{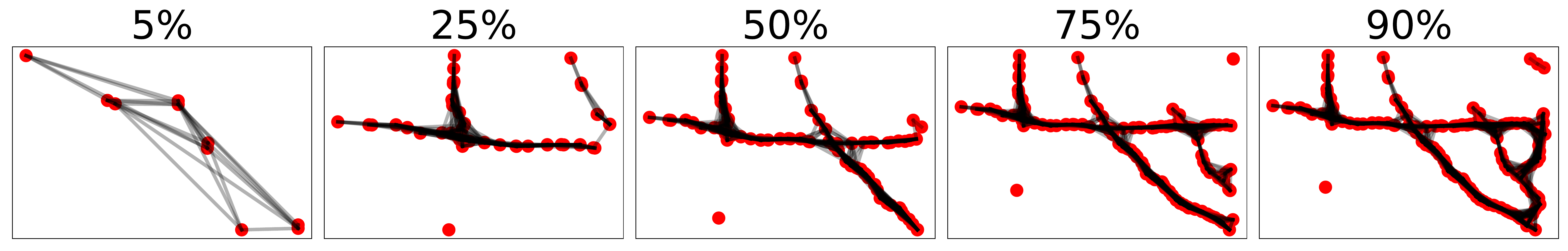}
        \caption{\update{METR-LA}}
    \end{subfigure}
    \caption{\update{Visualization of subgraphs visible in training under different ratios.}}
    \label{fig:vis-induc}
\end{figure}

\update{
We have further added results using 90\% and 5\% data on both datasets and we show the table of inductive learning results as \tref{tab:inductive}. We observe that: (1) With the portion of visible nodes in the training stage increasing, the prediction error of CNFGNN decreases drastically. However, the increase of the portion of visible nodes has negligible contribution to the performance of GRU + FedAvg after the portion surpasses 25\%. Since increasing the ratio of seen nodes in training introduces more complex relationships among nodes to the training data, the difference of performance illustrates that CNFGNN has a stronger capability of capturing complex spatial relationships. (2) When the ratio of visible nodes in training is extremely low (5\%), there is not enough spatial relationship information in the training data to train the GN module in CNFGNN, and the performance of CNFGNN may not be ideal. We visualize the subgraphs visible in training under different ratios in \fref{fig:vis-induc}.  However, as long as the training data covers a moderate portion of the spatial information of the whole graph, CNFGNN can still leverage the learned spatial connections among nodes effectively and outperforms GRU+FedAvg. We empirically show that the necessary ratio can vary for different datasets (25\% for PEMS-BAY and 50\% for METR-LA).
}

% \subsection{Ablation Study: Effect of Node-Level Model Aggregation}
% \begin{table}[htbp]
% \centering
% \begin{tabular}{@{}ccc@{}}
% \toprule
% Method & METR-LA & PEMS-BAY\\ \midrule
% GRU + \modelshortname~(w/o FedAvg) & &\\
% GRU + \modelshortname & 11.487 & 3.822 \\ \bottomrule
% \end{tabular}
% \caption{Effect of Node-Level Model Aggregation.}
% \end{table}

% \subsection{Ablation Study: Effect of Alternating Training of Node-Level and Spatial Models}
\subsection{\update{Ablation Study: Effect of Alternating Training and FedAvg on Node-Level and Spatial Models}}

\begin{figure}[t]
    \centering
    \begin{subfigure}[b]{0.9\linewidth}
        \centering
        \includegraphics[width=\textwidth]{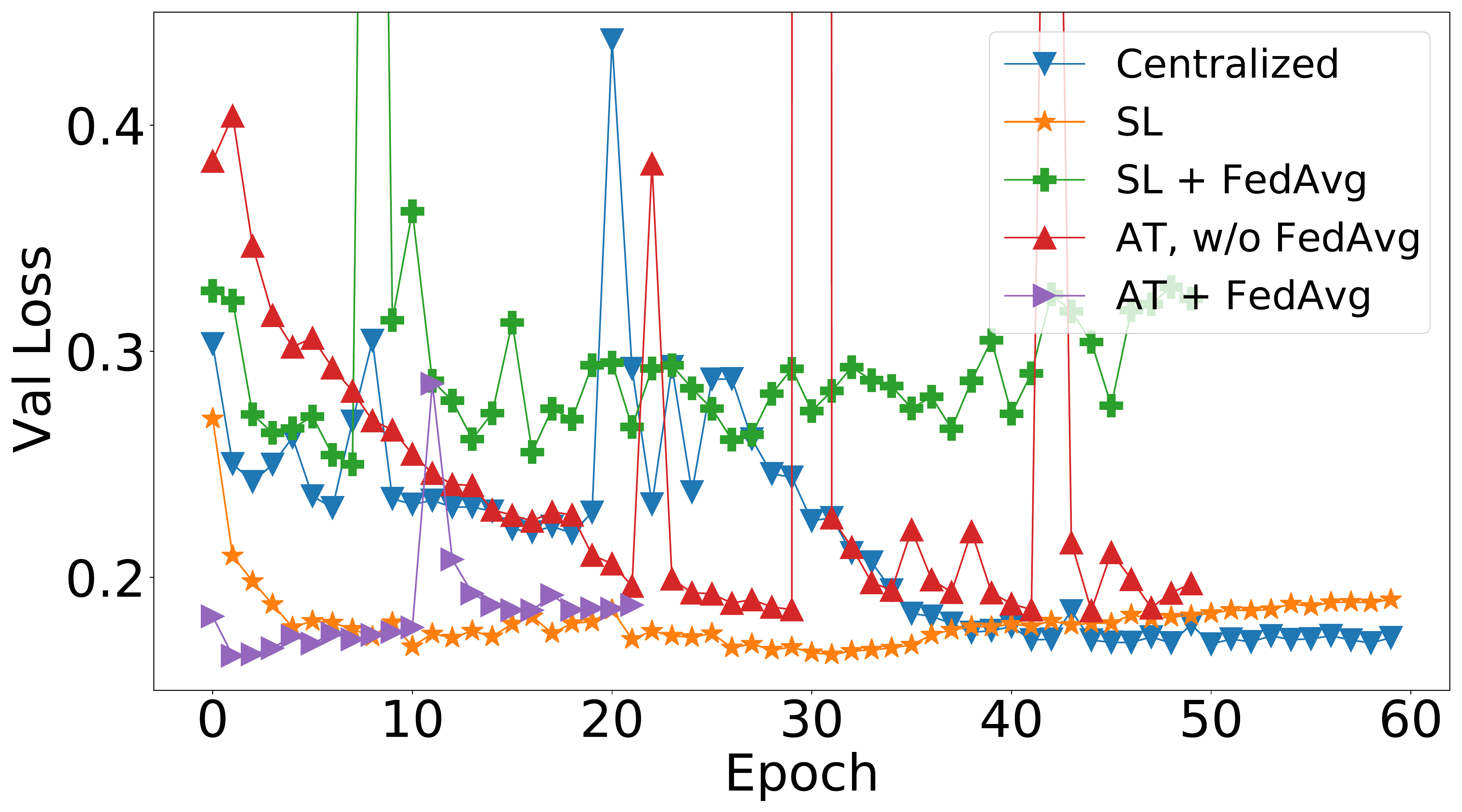}
        \caption{\update{PEMS-BAY}}
    \end{subfigure}
    \begin{subfigure}[b]{0.9\linewidth}
        \centering
        \includegraphics[width=\textwidth]{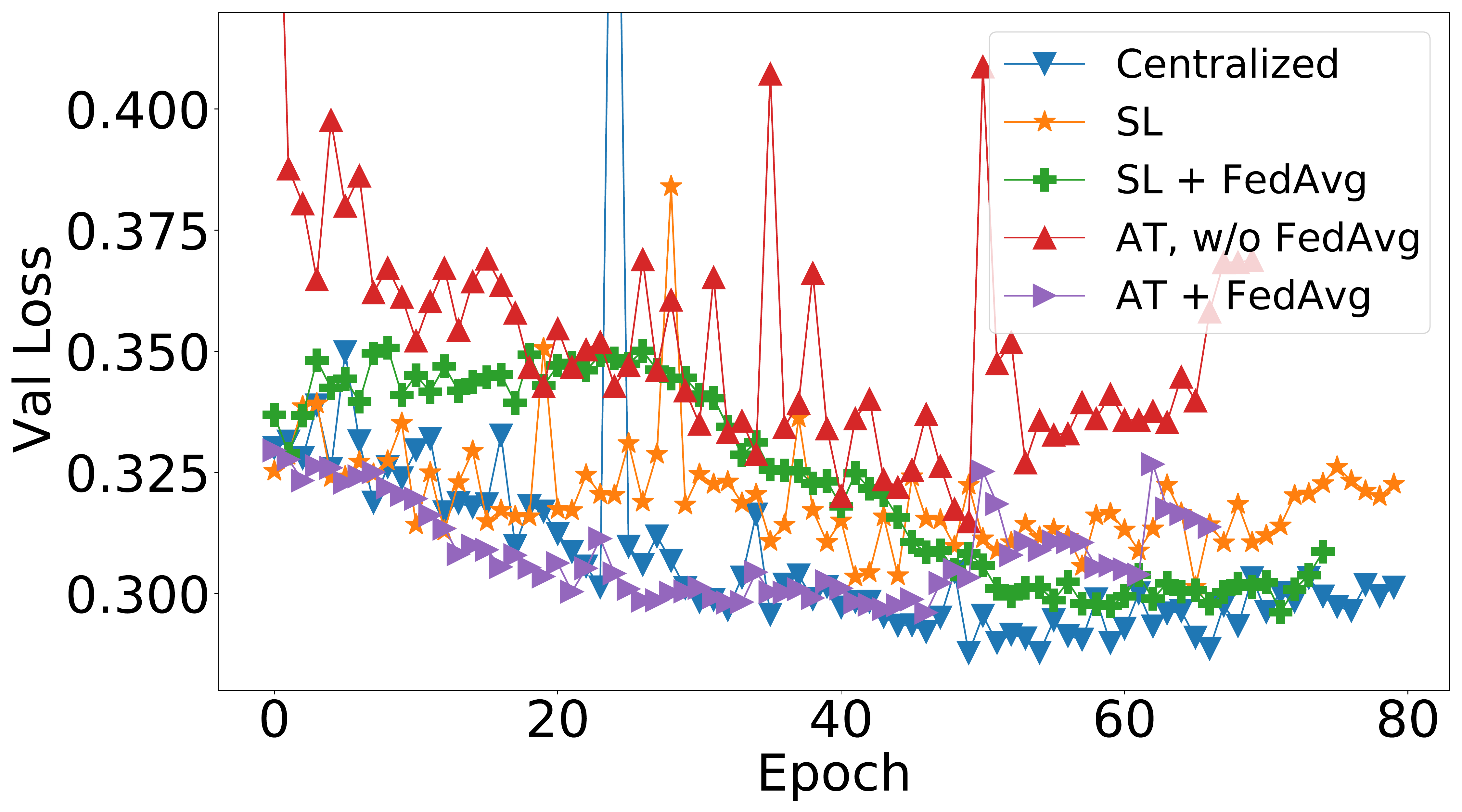}
        \caption{\update{METR-LA}}
    \end{subfigure}
    \caption{\update{Validation loss during the training stage of different training strategies.}}
    \label{fig:train-strat-val}
    % \vspace{-1em}
\end{figure}
% \paragraph{Baselines} We compare the effect of different training strategies of \modelshortname: (1) \textbf{Centralized}: \modelshortname~trained with centralized data where all nodes share one single encoder-decoder. (2) \textbf{Split Learning}: \modelshortname~trained with split learning~\citep{singh2019detailed}, where models on nodes and the model on the server are jointly trained by exchanging hidden vectors and gradients. (3) Alternating training without Federated Averaging of models on nodes (\textbf{AT, w/o FedAvg}). (4) Alternating training with Federated Averaging on nodes described in Section~\ref{subsubsec:alternating-training} (\textbf{AT + FedAvg}). 

% \begin{wraptable}{r}{0.55\textwidth}
% \centering
% % \vspace{-2pt}
% \caption{\update{Comparison of test error (RMSE) of different training strategies of CNFGNN.}}
% \label{tab:training-strategy}
% \begin{tabular}{@{}ccc@{}}
% \toprule
% \begin{tabular}[c]{@{}c@{}}Training\\ Strategy\end{tabular} & PEMS-BAY       & METR-LA         \\ \midrule
% Centralized                                                 & \textbf{3.816} & \textbf{11.471} \\ \midrule
% SL                                                          & 3.914          & 12.186          \\
% SL + FedAvg                                                 & 4.383          & 11.631          \\
% AT, w/o FedAvg                                              & 4.003          & 11.912          \\
% AT + FedAvg                                                 & \textbf{3.822} & \textbf{11.487} \\ \bottomrule
% \end{tabular}
% \end{wraptable}

\begin{table}[htbp]
\centering
% \vspace{-2pt}
\caption{\update{Comparison of test error (RMSE) and the communication cost during training of different training strategies of CNFGNN.}}
\label{tab:training-strategy}
\resizebox{\linewidth}{!}{%
\begin{tabular}{@{}ccccc@{}}
\toprule
\multirow{2}{*}{Method} & \multicolumn{2}{c}{PEMS-BAY}                                           & \multicolumn{2}{c}{METR-LA}                                             \\ \cmidrule(l){2-5}                      & RMSE  & \begin{tabular}[c]{@{}c@{}}Train Comm\\ Cost (GB)\end{tabular} & RMSE   & \begin{tabular}[c]{@{}c@{}}Train Comm\\ Cost (GB)\end{tabular} \\ \midrule
Centralized                                                 & \textbf{3.816} & -& \textbf{11.471}&- \\ \midrule
SL                                                          & 3.914&     350.366     & 12.186        & 307.627 \\
SL + FedAvg                                                 & 4.383 &     80.200    & 11.631        & 343.031 \\
AT, w/o FedAvg                                              & 4.003  &    5221.576    & 11.912        & 2434.985 \\
AT + FedAvg                                                 & \textbf{3.822}& 237.654 & \textbf{11.487}& 222.246 \\ \bottomrule
\end{tabular}
}
\end{table}

\paragraph{Baselines} We compare the effect of different training strategies of \modelshortname: (1) \textbf{Centralized}: \modelshortname~trained with centralized data where all nodes share one single encoder-decoder. (2) \textbf{Split Learning (SL)}: \modelshortname~trained with split learning~\citep{singh2019detailed}, where models on nodes and the model on the server are jointly trained by exchanging hidden vectors and gradients. \update{(3) \textbf{Split Learning + FedAvg (SL + FedAvg)}: A variant of SL that synchronizes the weights of encoder-decoder modules periodically with FedAvg.} (4) Alternating training without Federated Averaging of models on nodes (\textbf{AT, w/o FedAvg}). (5) Alternating training with Federated Averaging on nodes described in Section~\ref{subsubsec:alternating-training} (\textbf{AT + FedAvg}). 

% \paragraph{Discussion} 
% % \tref{tab:train-strat-test} shows that the adopted strategy AT + FedAvg achieves the lowest forecasting error on the METR-LA dataset.
% \fref{fig:train-strat-val} shows the effect of training strategies. We notice that (1) Split Learning quickly overfits the training data while AT w/o FedAvg and AT + FedAvg do not, which demonstrates that the alternating training helps the model avoid overfitting. (2) AT w/o FedAvg suffers from slow convergence and converges at suboptimal validation loss, while AT + FedAvg achieves the lowest validation error and performs closely to the model trained on centralized data. This result indicates that the federated learning of models on nodes is critical for \modelshortname~to achieve the best performance.

\paragraph{Discussion} 
\update{
% \fref{fig:train-strat-val} shows the effect of training strategies. We notice that (1) Split Learning quickly overfits the training data while AT w/o FedAvg and AT + FedAvg do not, which demonstrates that the alternating training helps the model avoid overfitting. (2) AT w/o FedAvg suffers from slow convergence and converges at suboptimal validation loss, while AT + FedAvg achieves the lowest validation error and performs closely to the model trained on centralized data. This result indicates that the federated learning of models on nodes is critical for \modelshortname~to achieve the best performance.
\fref{fig:train-strat-val} shows the validation loss during training of different training strategies on PEMS-BAY and METR-LA datasets, and \tref{tab:training-strategy} shows their prediction performance and the communication cost in training. We notice that (1) SL suffers from suboptimal prediction performance and high communication costs on both datasets; SL + FedAvg does not have consistent results on both datasets and its performance is always inferior to AT + FedAvg. AT + FedAvg consistently outperforms other baselines on both datasets, including its variant without FedAvg. (2) AT + FedAvg has the lowest communication cost on METR-LA and the 2nd lowest communication cost on PEMS-BAY, on which the baseline with the lowest communication cost (SL + FedAvg) has a much higher prediction error (4.383 vs 3.822). Both illustrate that our proposed training strategy, SL + FedAvg, achieves the best prediction performance as well as low communication cost compared to other baseline strategies.
}

\subsection{Ablation Study: Effect of Client Rounds and Server Rounds}
\label{subsec:client-server-rounds}
% \todo{results of different compositions of local rounds and server rounds}

% \begin{figure}[htbp]
% \centering
% \begin{minipage}{0.4\textwidth}
% \centering
% \includegraphics[width=\linewidth]{example-image-a}
% \caption{Forecasting error.}
% \end{minipage}
% \hspace{0.1\textwidth}
% \begin{minipage}{0.45\textwidth}
% \centering
% \includegraphics[width=\linewidth]{figures/rmse-comm-cost.pdf}
% \caption{Effect of local rounds ($R_c$) and server rounds ($R_s$) on forecasting performance and communication cost.}
% \end{minipage}
% \end{figure}

\paragraph{Set-up} We further investigate the effect of different compositions of the number of client rounds ($R_s$) in \aref{alg:client-training} and the number of server rounds ($R_c$) in \aref{alg:server-training}. To this end, we vary both $R_c$ and $R_s$ over [1,10,20]. 

\begin{figure}[t]
% \vspace{-2em}
  \begin{center}
    \includegraphics[width=0.9\linewidth]{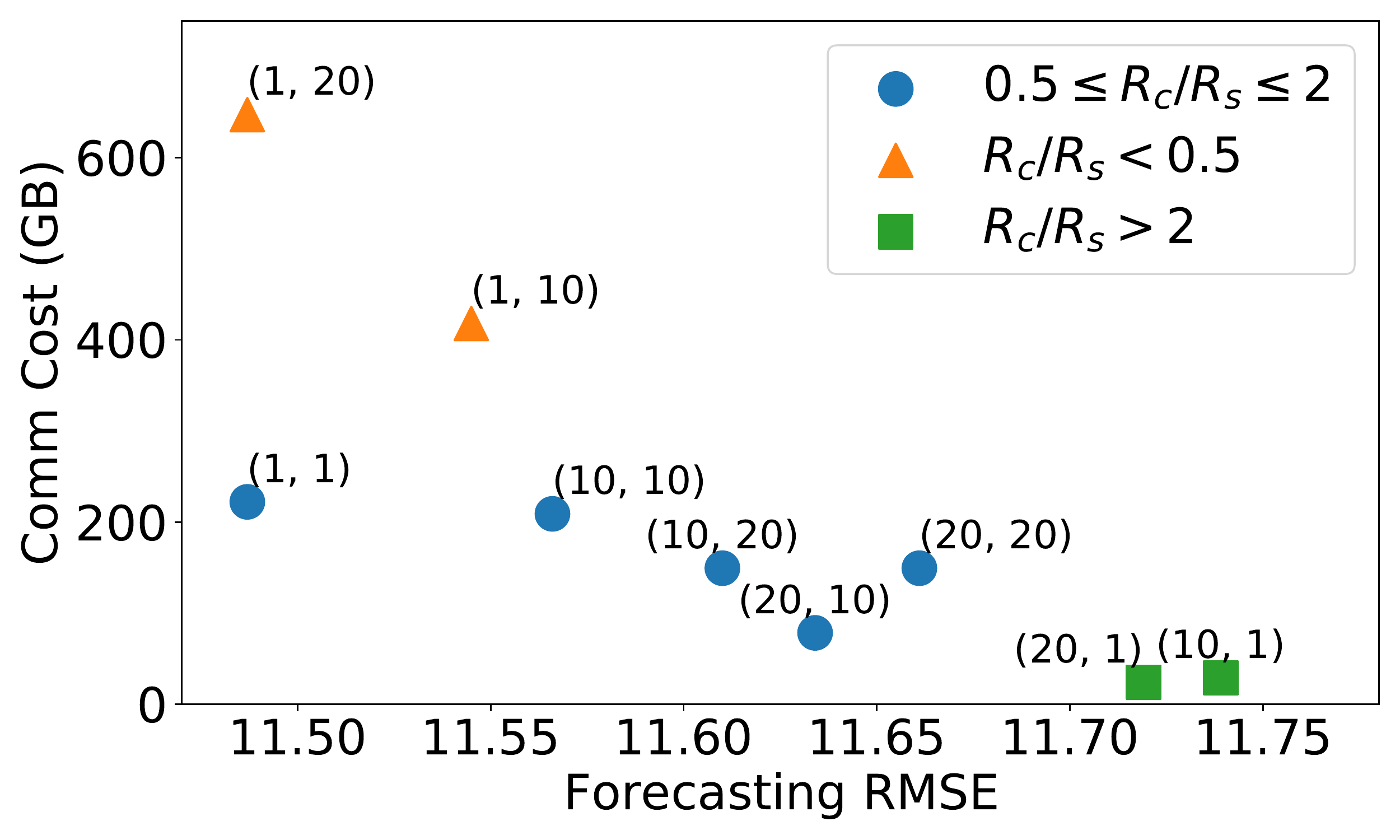}
  \end{center}
%   \vspace{-1em}
  \caption{Effect of client rounds and server rounds ($R_c$, $R_s$) on forecasting performance and communication cost.}
  \label{fig:local-server}
% \vspace{-1em}
\end{figure}

\paragraph{Discussion} \fref{fig:local-server} shows the forecasting performance (measured with RMSE) and the total communication cost in the training of \modelshortname~under all compositions of ($R_c$, $R_s$) on the METR-LA dataset. We observe that: (1) Models with lower $R_c/R_s$ ratios ($R_c/R_s<0.5$) tend to have lower forecasting errors while models with higher $R_c/R_s$ ratios ($R_c/R_s>2$) have lower communication cost in training. This is because the lower ratio of $R_c/R_s$ encourages more frequent exchange of node information at the expense of higher communication cost, while the higher ratio of $R_c/R_s$ acts in the opposite way. (2) Models with similar $R_c/R_s$ ratios have similar communication costs, while those with lower $R_c$ values perform better, corroborating our observation in (1) that frequent node information exchange improves the forecasting performance.

\section{Conclusion}
We propose \modelfullname~(\modelshortname), which bridges the gap between modeling complex spatio-temporal data and decentralized data processing by enabling the use of graph neural networks (GNNs) in the federated learning setting. We accomplish this by decoupling the learning of local temporal models and the server-side spatial model using alternating optimization of spatial and temporal modules based on split learning and federated averaging. Our experimental results on traffic flow prediction on two real-world datasets show superior performance as compared to competing techniques. Our future work includes \update{applying existing GNN models with sampling strategies and integrating them into \modelshortname~for large-scale graphs, }extending \modelshortname~to a fully decentralized framework, and incorporating existing privacy-preserving methods for graph learning to \modelshortname, to enhance federated learning of spatio-temporal dynamics.

\begin{acks}
This work is supported in part by NSF Research Grant IIS-1254206, NSF Research Grant CCF-1837131, and WeWork, granted to co-author Yan Liu in her academic role at the University of Southern California.
The views and conclusions are those of the authors and should not be interpreted as representing the official policies of the funding agency, or the U.S. Government.
\end{acks}

\bibliographystyle{ACM-Reference-Format}
\bibliography{references}

\newpage
\appendix
\setcounter{table}{0}
\renewcommand{\thetable}{A\arabic{table}}
\setcounter{figure}{0}
\renewcommand{\thefigure}{A\arabic{figure}}
\setcounter{algorithm}{0}
\renewcommand{\thealgorithm}{A\arabic{algorithm}}
\setcounter{equation}{0}
\renewcommand{\theequation}{A\arabic{equation}}
\section{Appendix}
\subsection{\update{The Histograms of Data on Different Nodes}}

\begin{figure}[htbp]
    \centering
    \includegraphics[width=\linewidth]{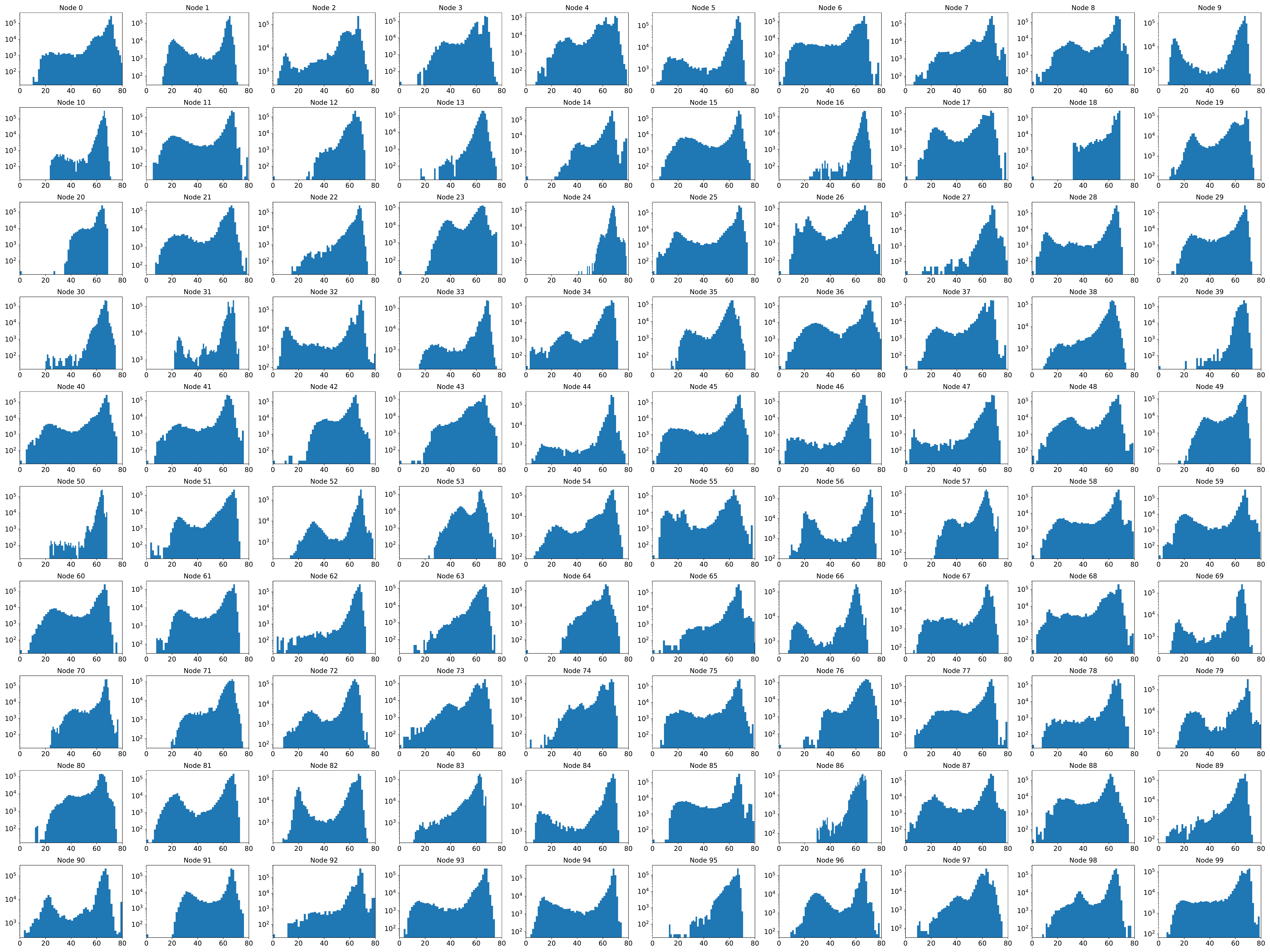}
    \caption{\update{The histograms of PEMS-BAY data on the first 100 nodes ranked by ID.}}
    \label{fig:hist-nodes-pems-bay}
\end{figure}

\begin{figure}[htbp]
    \centering
    \includegraphics[width=\linewidth]{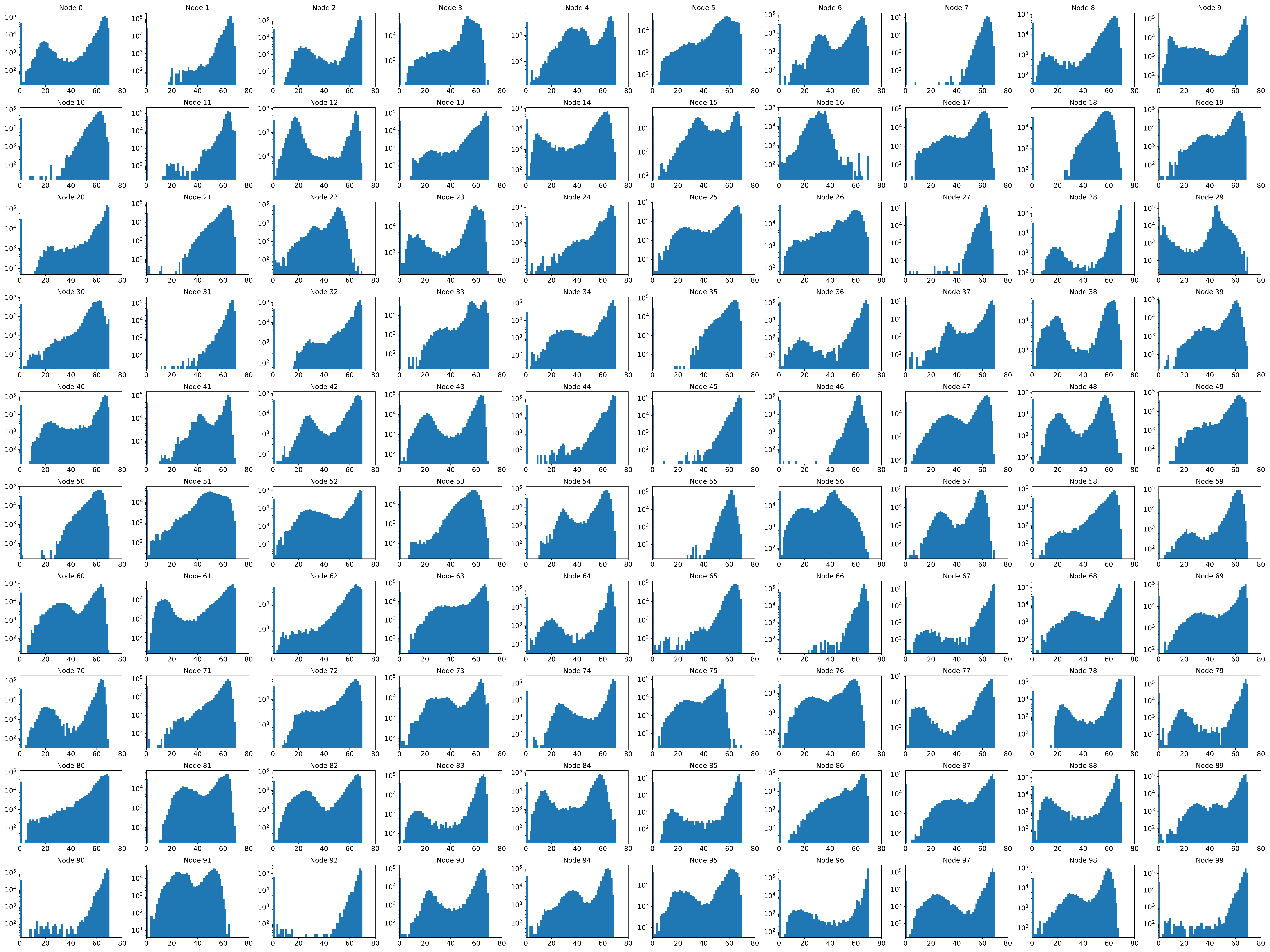}
    \caption{\update{The histograms of METR-LA data on the first 100 nodes ranked by ID.}}
    \label{fig:hist-nodes-metr-la}
\end{figure}

\update{
We show the histograms of traffic speed on different nodes of PEMS-BAY and METR-LA in \fref{fig:hist-nodes-pems-bay} and \fref{fig:hist-nodes-metr-la}. For each dataset, we only show the first 100 nodes ranked by their IDs for simplicity. The histograms show that the data distribution varies with nodes, thus data on different nodes are not independent and identically distributed.
}

% \begin{figure*}[htbp]
%     \centering
%     \begin{subfigure}[b]{\textwidth}
%         \centering
%         \includegraphics[width=\textwidth]{figures/pems-bay-value-hist.pdf}
%         \caption{\update{PEMS-BAY}}
%     \end{subfigure}
%     \begin{subfigure}[b]{\textwidth}
%         \centering
%         \includegraphics[width=\textwidth]{figures/metr-la-value-hist.pdf}
%         \caption{\update{METR-LA}}
%     \end{subfigure}
%     \caption{\update{The histograms of data on the first 100 nodes ranked by ID.}}
%     \label{fig:hist-nodes}
% \end{figure*}

\subsection{Table of Notations}
Table~\ref{tab:notations} summarizes notations used in this paper and their definitions.
\begin{table}[H]
    \centering
    \caption{Table of notations.}
    \label{tab:notations}
    \begin{tabular}{cl}
    \toprule
        Notation & Definition \\ \midrule
        $\mathcal{G} = (\mathcal{V}, \mathcal{E})$ & \begin{tabular}{@{}l@{}}Graph $\mathcal{G}$ defined with the set of nodes $\mathcal{V}$\\and the set of edges $\mathcal{E}$.\end{tabular}\\
        $\tX$ & Tensor of node features. $\tX\in\mathbb{R}^{\left\lvert \mathcal{V} \right\rvert \times \dots}$.\\
        $\vx_i$ & Features of the $i$-th node.\\
        $\tY$ & Tensor of node labels for the task. $\tY\in\mathbb{R}^{\left\lvert \mathcal{V} \right\rvert \times \dots}$.\\
        $\vy_i$ & Labels of the $i$-th node.\\
        $\hat{\vy}_i$ & Model prediction output for the $i$-th node.\\
        $R_g/R_c/R_s$ & \begin{tabular}{@{}l@{}}Maximum number of global/server/client\\training rounds.\end{tabular}\\
        $\vtheta_{GN}^{(r_g)}$ & \begin{tabular}{@{}l@{}}Weights of the server-side Graph Network\\in the $r_g$-th global training round.\end{tabular}\\
        $\bar{\vtheta}^{(r_g)}_{c}$ & \begin{tabular}{@{}l@{}}Aggregated weights of client models\\in the $r_g$-th global training round.\end{tabular}\\
        $\vh_{c,i}$ & \begin{tabular}{@{}l@{}}Local embedding of the input sequence\\of the $i$-th node.\end{tabular} \\
        $\vh_{G,c,i}$ & \begin{tabular}{@{}l@{}}Embedding of the $i$-th node propagated\\with the server-side GN.\end{tabular} \\
        $\ell_i$ & \begin{tabular}{@{}l@{}}Loss calculated on the $i$-th node.\end{tabular} \\
        $\eta_s$ & Learning rate for training $\vtheta_{GN}^{(r_g)}$.\\
        $\eta_c$ & Learning rate for training $\bar{\vtheta}^{(r_g)}_{c}$.\\
        \bottomrule
    \end{tabular}
\end{table}

\end{document}